\pdfoutput=1
\documentclass[11pt]{article}

\usepackage[final]{acl}

\usepackage{times}
\usepackage{latexsym}
\usepackage{threeparttable}

\usepackage[T1]{fontenc}
\usepackage[utf8]{inputenc}

\usepackage{microtype}

\usepackage{inconsolata}

\usepackage{graphicx}
\usepackage{multirow}

\usepackage{algorithm}
\usepackage{algpseudocode}
\usepackage{amsmath}
\usepackage{amssymb}
\title{Harmonizing Diverse Models: A Layer-wise Merging Strategy for Consistent Generation}

\author{
  Xujun Peng, Anoop Kumar, Jingyu Wu, Parker Glenn, Daben Liu \\
  AI Foundations, Capital One \\ McLean, VA, USA \\
  \{xujun.peng, anoop.kumar, jingyu.wu, parker.glenn, daben.liu\}@capitalone.com
}

\begin{document}
\maketitle

\begin{abstract}
%Large Language Models (LLMs) frequently generate inconsistent outputs when exposed to semantically equivalent but varied inputs, posing a significant challenge to their reliability and deployability, a problem exacerbated by the scarcity of dedicated consistency-focused datasets. Current fine-tuning methods often fail to address this issue in a robust way. To overcome these limitations, we propose a comprehensive methodology: we systematically generate diverse synthetic datasets, integrate Triplet Loss for enhanced embedding learning, and introduce a novel layer-wise model merging approach. This merging strategy utilizes consistency-oriented weights derived from an in-depth analysis of individual layer activations, thereby consolidating knowledge from models specialized in handling different input variations. Results show the proposed method significantly improves consistency measurements compared to baselines and individual models, demonstrating a practical approach for building more reliable RAG generators.

%%Large Language Models (LLMs) often generate inconsistent outputs for semantically equivalent inputs, a challenge worsened by scarce consistency data and ineffective fine-tuning. We introduce a methodology featuring systematic synthetic data generation, Triplet Loss for better embeddings, and a novel layer-wise model merging approach. This strategy employs consistency-oriented weights derived from layer activations to consolidate knowledge from specialized models. Our method significantly boosts consistency, providing a practical solution for reliable RAG generators.

Retrieval-Augmented Generation (RAG) systems  leverage Large Language Models (LLMs) to generate accurate and reliable responses that are grounded in retrieved context. However, LLMs often generate inconsistent outputs for semantically equivalent inputs, a problem compounded by the scarcity of consistency-focused training data and the limitations of current fine-tuning techniques in enhancing output consistency. We propose a new approach combining systematic synthetic data generation, triplet loss for better embeddings, and a novel layer-wise model merging approach. Using consistency-aware weights derived from intermediate layer activations, our method effectively integrates knowledge from specialized models. Experimental results how that our merged model significantly enhances output consistency, achieving  a ~47.5\% improvement in response  similarity  over the baseline, thus offering a practical solution for increasing the reliability of an industrial RAG system.

\end{abstract}

\section{Introduction}

LLMs have demonstrated remarkable capabilities in natural language understanding and generation,  enabling breakthroughs across a broad spectrum of applications such as question answering and summarization. RAG has emerged as a powerful paradigm that combines the generative strength of LLMs with external knowledge retrieval to enhance factuality, reduce hallucination, and extend context beyond model limitations \cite{lewis2020retrieval, wu2024retrieval}. 

\begin{figure}[hpt]
    \centering
    \includegraphics[width=\columnwidth, trim=2.5cm 5.6cm 2.5cm 2.5cm, clip]{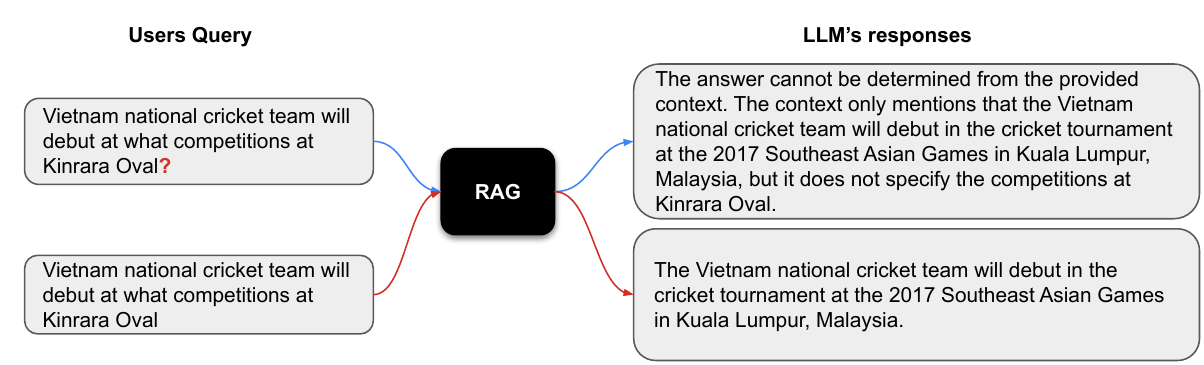}
    \caption{Variability in LLM Responses from Subtle Query Differences.}
    \label{fig:query}
\end{figure}

Despite their potential, RAG systems often generate inconsistent responses, for minor and semantically insignificant variations in the input query or the prompt \cite{song2024survey}. This inconsistency manifests itself in various forms, including contradictory responses, variability in factual grounding, and fluctuations in the level of detail or confidence expressed by the model. This unpredictability not only undermines the reliability of RAG systems but also poses challenges for their adoption in high-stakes or knowledge-sensitive domains such as finance, healthcare, and scientific research.
As shown in Figure \ref{fig:query}, the mere presence or absence of a question mark can dramatically alter the response of a RAG based QA system. 
In an industrial production deployment, there could several such variations in how users query the system, posing challenges in the adoption of RAG systems.

In RAG systems, two key models work together: the retriever and the generator. The retriever is responsible for fetching relevant content based on a user query or prompt, while the generator creates a coherent and contextually appropriate answer by leveraging both the query and the retrieved content. 

Inconsistencies may arise during either the retrieval or the generation process, leading to varied responses.  However, our empirical observations and \citet{zuccon2016query},  \citet{abdallah2025retrieval} indicate that generators are more sensitive and less consistent than retrievers to minor variations in queries. While retrievers tend to be consistent, even in the face of minor, semantically insignificant variations in the input query (e.g., different phrasings of the same question), generators exhibit higher variability \cite{cao2025out}. Small changes in phrasing or query structure can lead to different answers being generated, even when the retrieved content remains the same. This difference in behavior highlights the challenges faced by generative models in maintaining consistency, especially in tasks where precise and coherent responses are required. \par

In this work, we focus on characterizing, measuring, and mitigating such inconsistency. Our main contributions are as follows: 
\begin{enumerate}
\item We characterize different types of query variations that lead to inconsistent responses  in an industrial RAG system.
\item We identify metrics and demonstrate inconsistency in the question answering task.
\item We develop novel layer-wise merging approach to reduce inconsistency without regressing on accuracy. 
\end{enumerate}

%%Despite their remarkable fluency in tasks such as question answering and text summarization, Large Language Models (LLMs) face a critical, persistent challenge: their inconsistent responses, even when presented with semantically equivalent inputs. Current research highlights that these models are acutely sensitive to subtle prompt variations. For instance, as depicted in Figure \ref{fig:query}, the mere presence or absence of a question mark can dramatically alter outputs from a Retrieval Augmented Generation (RAG)-based QA system. Maintaining strict semantic consistency across such inputs is paramount for the reliability of many language generation applications. This issue becomes particularly critical in high-stakes environments like the financial industry, where inconsistent or inaccurate answers to similar inquiries could lead to severe legal ramifications, thus posing a substantial obstacle to trusted deployment.

\section{Related Work}
We review the literature to establish a clear understanding of what constitutes consistency, how it is measured, and strategies to improve consistency in context of LLMs. A widely accepted concept, as proposed by \citet{10835426}, defines consistency as the semantic equivalence of two responses generated by the same LLM when prompted with identical or semantically similar queries multiple times. \par

%Given the scarcity of resources and datasets for improving/evaluating LLM consistency, researchers have increasingly focused on constructing specialized datasets for this purpose. For instance, \cite{elazar-etal-2021-measuring} introduced a high-quality resource for measuring factual consistency in pre-trained LLMs and developed a novel consistency loss method that effectively enhances their consistency, even on unseen relations. Similarly, \cite{raj2025semanticconsistencyassuringreliability} proposed and released PromptSET, a new dataset built upon TriviaQA and HotpotQA, specifically designed to investigate the impact of subtle prompt variations on LLM behavior. Their findings underscore that current methods struggle to predict prompt sensitivity effectively, emphasizing the need to understand optimal prompt phrasing for consistent and accurate LLM responses. In \cite{zhao2024consistencymattersexplorellms}, a method of developing an automated evaluation tool for LLM consistency, built on a custom dataset and utilizing LightGBM with traditional Natural Language Generation (NLG) metrics, was proposed. In targeting the inconsistency of LLMs in generating natural-language explanations across related inputs, explanation-consistency finetuning (EC-finetuning) was proposed, which adapts LLMs to produce more consistent explanations by training them on carefully constructed synthetic data \cite{chen-etal-2025-towards}. 

Evaluating and improving LLM consistency remains challenging due to a lack of targeted benchmarks and datasets, prompting research into specialized evaluation methods. \citet{elazar-etal-2021-measuring} contributed a valuable resource for factual consistency measurement in pre-trained LLMs and a novel consistency loss to enhance performance even on new data. To explore prompt sensitivity, \citet{raj2025semanticconsistencyassuringreliability} introduced PromptSET, \citet{qiang2024prompt}  applied perturbations to highlight the difficulties in predicting how minor prompt changes affect LLMs. For evaluation, \citet{zhao2024consistencymattersexplorellms}  propose an automated consistency assessment tool using a custom data set. Addressing inconsistency in natural-language explanations, \citet{chen-etal-2025-towards} developed EC-finetuning, that trains  on synthetic data to  increase consistency. 

%%%Addressing the challenge of LLM consistency is hindered not only by the limited resources available for dedicated consistency training and evaluation, but also by the absence of effective metrics. Conventional accuracy-based measures, such as ROUGE and BLEU, often prove inadequate in capturing the nuanced inconsistencies inherent in LLM outputs. \par

To address the need for reliable LLM output in NLG, especially under semantically equivalent inputs, \citet{raj2023measuringreliabilitylargelanguage} proposed a framework to measure semantic consistency via output agreement, showing strong correlation with human judgments across domains. Complementing this, \citet{wu2025estimatingllmconsistencyuser} introduced a logit-based ensemble method aligned with human perceptions through a user study. \citet{lee-etal-2025-evaluating} examined LLM consistency as automated evaluators, focusing on their reliability in scoring identical items.

Recent work has focused on actively improving LLM consistency. \citet{raj2025improvingconsistencylargelanguage} used multi-step prompting and synthetic data for semantic alignment. \citet{sathe-etal-2025-improving} enhanced self-consistency via multiple input tokenizations. \citet{wang2023selfconsistencyimproveschainthought} improved reasoning by sampling diverse outputs and selecting by voting.

\section{Methodology}
%In this section, we detail our methodology for improving the consistency of LLM outputs, specifically within the context of a Retrieval-Augmented Generation (RAG) system.

In this section, we describe our methodology for improving the consistency of LLM responses, specifically within the context of a RAG system. 

We observe that the retriever in our RAG system provides accurate context, but minor variations in the query often lead to inconsistent response from the generator. This work focuses on improving generator consistency under such variations, assuming stable retrieval quality. Addressing retrieval inconsistency is left for future work. 

Our work aims to improve the RAG generator's consistency. By analyzing human-annotated data, we constructed diverse synthetic datasets to train multiple individual generator models. To achieve higher response consistency, we developed a novel consistency-focused, layer-wise model merging approach, building upon DARE-TIES \cite{10.5555/3692070.3694452, 10.5555/3666122.3666432}. This strategy allowed us to effectively combine knowledge from individual models trained on diverse synthetic data.

\subsection{Synthetic Data Generation} \label{sec:synthetic}
% One seemingly simple way to make LLMs more consistent is to include every conceivable input variation in their training data, then train or fine-tune the models to recognize these variations. However, this approach often isn't practical because acquiring such comprehensive data is extremely difficult, especially in many industry sectors. Consider the challenges in healthcare: data remains isolated in different hospitals, and the immense volume and diverse nature of medical terminology make it nearly impossible to compile truly exhaustive training data. The financial industry faces a similar hurdle, where strict privacy regulations severely restrict the ability to collect sufficient data with the necessary variations. \par

A direct approach to improving LLM consistency is to train on all possible input variations. However, this is impractical due to the vast number of potential variations and limited data availability - especially in domains like healthcare, where data is fragmented and complex, and finance, where privacy regulations constrain access. \par 

\begin{table*}[htp]
    \centering
    \small
    %\caption{Queries in live data contributing variations that leads to different answers 
    %(Domain Information Replaced with Analogous General Content).}
    \caption{Illustrative Query Variations Leading to Different Answers.}
    \begin{tabular}{c| p{0.3\linewidth} | p{0.3\linewidth}}
        \hline
        Variation Type & Query & Query'  \\ \hline\hline
        %Omissions & \textbf{how to} clean garden tools at end of season & clean garden tools at end of season \\
        Variation - How do/to & \textbf{how do we} manage customer feedback at end of project & \textbf{how to} manage customer feedback at end of project  \\
        I vs. we & can \textbf{we} drive to a grocery store & can \textbf{I} drive to a grocery store  \\
        Singular vs. plural & delivering \textbf{packages} for shipment & delivering \textbf{package} for shipment  \\
        Article omissions & how to add \textbf{a contact} to \textbf{a} phone book & how to add \textbf{contacts} to phone \textbf{books}  \\ \hline
    \end{tabular}   
    \label{tab:variation_stat}
\end{table*}

% Given these limitations, training models with generated synthetic data offers a feasible and widely used method to improve generalizability and consistency with limited real-world data. Through our analysis of the live queries we've gathered, we discovered that a few key categories of inquiries are responsible for most of the input diversity in our RAG system. The table \ref{tab:variation_stat} below details each query type, which typically feature semantically identical but syntactically varied inputs, and provides illustrative examples.

% Given these limitations, synthetic data offers a practical means to improve consistency when real data is scarce. Our analysis of production queries shows that a small set of key variations accounts for the majority of input diversity in  industrial RAG systems.  Table \ref{tab:variation_stat} lists the types of variations we observe with examples of semantically similar but syntactically different inputs.These variations differ from public benchmarks that capture typos, keyboard proximity errors, etc. \cite{zhang2025qe}.

Given these limitations, synthetic data provides a practical way to improve consistency when real data is scarce. While prior work has documented a broad range of general query variations - such as typos, synonyms, keyboard proximity errors, and paraphrases \cite{zhang2025qe, wu2025estimatingllmconsistencyuser} - it does not capture several nuanced variations observed in large-scale industrial RAG systems. Our analysis of production queries shows that a small set of key variations accounts for most input diversity, often involving subtle rephrasings (e.g., ``how to we manage an account'' vs. ``how to manage an account'') rather than simple surface-level errors. Table \ref{tab:variation_stat} lists these variation types with representative examples. Characterizing and accounting for such variations is critical to improving system robustness and building user trust in real-world RAG applications.

Based on the analysis of our dataset, we've identified three main types of query variations: \par

\textbf{How to/do variations}: These queries often involve rephrasing questions about methods or actions. We used regular expression rules to systematically create additional queries of this nature. \par
%\begin{align*}
%    &\texttt{how (can|do) (i|we|you|customers|a customer|the customer)} \\
%    &\rightarrow \texttt{how to} \\
%    & \texttt{can (we|you|the customer|a customer|customers)} \\
%    & \rightarrow \texttt{can i}
%\end{align*}
\textbf{Singular/Plural/Article variations}: This category covers changes in noun quantity (e.g., ``apple'' vs. ``apples'') and the use of articles (e.g., ``a'', ``an'', ``the''). To synthesize more of these variations, we randomly interchanged singular and plural forms and substituted or modified articles.  \par

\textbf{Semantic variations}: These are changes in wording that maintain the same core meaning but use different vocabulary or phrasing. For semantic variations, we leveraged a pretrained LLM (Llama-3.1-70B-Instruct) to paraphrase our queries \cite{grattafiori2024llama3herdmodels}. \par

We used these synthetic queries to run our IR system, capturing updated contexts for our RAG system. This process generated enriched training and test datasets with a wide array of input variations. Rather than training/fine-tuning a single LLM with all the real-world and synthetic data, we opted to train/fine-tune multiple specialized models, each focusing on a different category of input variations. This approach allows each model to excel at the specific underlying tasks associated with its particular query type.

\subsection{Triplet Loss Training} \label{subsec:triplet}
Unlike traditional LLM fine-tuning that relies solely on cross-entropy loss, we incorporate triplet loss during our fine-tuning phase. \par

Triplet Loss \cite{schroff2015facenet} is a widely used loss function in metric learning, used in tasks such as face recognition and semantic search, to learn embeddings that pull similar items closer while pushing dissimilar ones apart. The core idea of Triplet Loss is to train on triplets of data points: an anchor $A$, a positive 
$P$ that is similar to the anchor, and a negative $N$  that is dissimilar. The objective is to ensure that the distance between $A$ 
and $P$ is smaller than that between $A$  and $N$. The triplet Loss function is formulated as:
\begin{equation}
    \begin{split}
        L(A, P, N) &= \max(0, d(f(A), f(P)) \\
        &\quad - d(f(A), f(N)) + \alpha)
    \end{split}
\end{equation}
More details of triplet loss can be found in \cite{7298682, reimers-gurevych-2019-sentence}. \par

In our implementation, triplets were constructed by first choosing an anchor query ($A$). We then selected its corresponding positive ($P$) and negative ($N$) data points by randomly sampling from its top 10 and bottom 10 nearest neighbors, respectively, within the feature space generated by a semantic feature extractor. \par

The final loss function employed during our training and fine-tuning process is a combination of cross-entropy loss and triplet loss, defined as:
\begin{equation}
    \mathcal{L} = \mathcal{L}_{\text{CE}} + \alpha \cdot \mathcal{L}_{\text{Triplet}},
\end{equation}
where $\alpha$ is a predefined weighting factor designed to balance the contribution of triplet loss.

\subsection{Model Merging} \label{subsec:merging}
With a suite of specialized models, each trained on distinct synthetic datasets, the challenge became generating a single, consistent response without sacrificing accuracy. While conventional ensemble approaches for multiple pre-trained or fine-tuned LLMs involve parallel execution and output combination, they incur significant computational costs  and inference latency  \cite{chen2025harnessing}. \par

To address these limitations, model merging offers a solution by consolidating knowledge from multiple pre-trained or fine-tuned models into  a single consolidated model. These techniques range from simple averaging to complex algorithms that align features and selectively transfer knowledge. Here, we introduce a novel model merging approach, building on the DARE-TIES merge method \cite{10.5555/3692070.3694452, 10.5555/3666122.3666432}, with the main goal of substantially boosting the consistency of the unified model's responses. \par

DARE-TIES merging is a sophisticated model merging algorithm designed to overcome the limitations of simple weight averaging, especially when combining fine-tuned models that originate from a common pre-trained base model but have diverged during training on different tasks or datasets. It operates on the principle of merging the $\Delta \theta_k = \theta_{F_k} - \theta_P $ that fine-tuned models apply to a common pre-trained model, rather than directly merging the absolute weights, where $\theta_P$ is the base model's parameters and $\theta_{F_k}$ denotes the parameters of the $k$-th fine-tuned model. By applying sparsification, sign matching and inverse-scaling on the $\Delta \theta_k$, DARE-TIES yields the merged model's parameters by: 
\begin{equation}
    \theta_{\text{merged}} = \theta_P + \sum_{k=1}^N \Delta \theta_k.
    \label{eq:merge_dare_ties}
\end{equation}

% To enhance the consistency of the model's responses for semantically identical inputs, we analyzed the consistency capabilities of each layer within our multiple trained LLMs and subsequently assigned varying weights in Equation \ref{eq:merge_dare_ties} for model merging. \par

To improve consistency with semantically identical inputs, we analyzed the consistency of each LLM layer, then assigned dynamic weights in Equation \ref{eq:merge_dare_ties} for merging.

To accomplish this, we first formed a development set $\mathcal{S}_{dev}$ of $T$ diverse data points. Then, for each model $k$ and each layer $l$, we extracted the activations $\alpha^{(l)}_k \in \mathbb{R}^{D \times T}$ from development set $\mathcal{S}_{dev}$, where $D$ represents the output feature dimension of layer $l$. For sequential outputs, we used max-pooling to extract these activations. This process enabled us to compute a similarity matrix $\Sigma^{(l)}_k \in \mathbb{R}^{T \times T}$ for the activations of each data point at every layer of model $k$. \par

Ideally, a model exhibiting high consistency with semantically identical inputs should produce similar activations within a single layer. Conversely, if inputs are semantically distinct, their activations should diverge significantly. Therefore, a consistent model would ideally yield similar similarity matrices $\Sigma^{(l)}_k$ across different layers when presented with the same set of inputs. \par

Leveraging this intuition, we can quantify a model's consistency by comparing the $\Sigma^{(l)}_k$ from different layers. Our approach begins by using a semantic feature extractor (specifically, a sentence transformer) to obtain features for each query in our development set, $\mathcal{S}_{dev}$. From these features, we computed a reference similarity matrix $\Sigma_r$. Subsequently, we quantified the discrepancy between each layer's similarity matrix $\Sigma^{(l)}_k$ and this reference using the absolute difference: $d^{(l)}_k = |\Sigma^{(l)}_k - \Sigma_r|$. 
Hence, for a specific layer $l$ across our various LLMs, we obtain a set of distance values, $\text{DM}^{(l)} = [d^{(l)}_1, d^{(l)}_2, \ldots, d^{(l)}_N]$. To convert these distances into weights that indicate a layer's contribution to consistency, we apply an inverted non-linear normalization approach. First, we computed the inverted distance for each layer's distance $d^{(l)}_k$ by subtracting it from the maximum distance observed for that layer across all models:
$$ \tilde{d}^{(l)}_k = \max(\text{DM}^{(l)}) - d^{(l)}_k $$
Next, these inverted distances are normalized to obtain $r^{(l)}_k$:
$$ r^{(l)}_k = \frac{\tilde{d}^{(l)}_k}{\sum_{j=1}^N \tilde{d}^{(l)}_j} $$
Finally, we apply a sigmoid function to these normalized inverted distances to derive the final weight $w^{(l)}_k$ for layer $l$ of model $k$:
\begin{equation}
    w^{(l)}_k = \sigma(a \cdot r^{(l)}_k + b)
\end{equation}
Here, $\sigma(\cdot)$ denotes the sigmoid function, and $a$ and $b$ are predefined scaling and offset parameters. \par

Based on the derived consistency-oriented layer weights $w^{(l)}_k$ for each model $k$, we modified Equation \ref{eq:merge_dare_ties} to incorporate these weights into the layer-wise model merging process:
\begin{equation}
    \theta^{(l)}_{\text{merged}} = \theta^{(l)}_P + \sum_{k=1}^N w^{(l)}_k \cdot \Delta \theta^{(l)}_k.
    \label{eq:updated_merge_dare_ties}
\end{equation}

The final merged LLM is constructed by applying Equation \ref{eq:updated_merge_dare_ties} in a layer-wise manner.

We outline the algorithm for  model merging:

\begin{algorithm}[H]
\caption{Consistency-Aware Model Merging}
\begin{algorithmic}[1]
\State \textbf{Input:} Base model $\theta_P$, fine-tuned models $\{\theta_{F_k}\}_{k=1}^N$, dev set $\mathcal{S}_{dev}$
\State \textbf{Output:} Merged model $\theta_{\text{merged}}$
\vspace{0.5em}

\State Compute reference similarity matrix $\Sigma_r$ using a sentence encoder on $\mathcal{S}_{dev}$

\For{each model $k$ and layer $l$}
    \State Extract activations and compute similarity matrix $\Sigma_k^{(l)}$
    \State Compute distance $d_k^{(l)} = |\Sigma_k^{(l)} - \Sigma_r|$
\EndFor

\For{each layer $l$}
    \State Normalize distances $d_k^{(l)}$ to weights $w_k^{(l)}$ using inverted scaling and sigmoid
    \State Merge: $\theta_{\text{merged}}^{(l)} = \theta_P^{(l)} + \sum_k w_k^{(l)} \cdot (\theta_{F_k}^{(l)} - \theta_P^{(l)})$
\EndFor

\State \Return $\theta_{\text{merged}}$
\end{algorithmic}
\end{algorithm}

\section{Experiments}
Our experimental setup utilized a QA engine built on a RAG architecture. For the evaluation of our consistency improvement method, the retriever component was held constant, and the generator component underwent fine-tuning.

\subsection{Datasets} \label{subsec:datasets}
% To fine-tune and evaluate our LLM-based generator, we built our training and test datasets using live operational data. We started by gathering 2,738 frequently used queries from our customer agents. Next, we used our production information retrieval (IR) system to fetch the relevant context for each of these queries. These 2,738 queries and their corresponding retrieved contexts then served as the inputs for our LLM generator. Domain experts annotated the expected answers for all these query-context pairs. Finally, we split this data into training and test subsets, with 1,421 for training and the remaining 1,317 for testing. \par

To fine-tune and evaluate our LLM generator, we used 2,738 representative queries and their retrieved contexts  that resemble a production IR system. Domain experts annotated the expected answers, and the data was split into 1,421 training and 1,317 test samples. \par

To get more varied inputs for training our model, we applied the methods detailed in Section \ref{sec:synthetic} to create three distinct types of synthetic data. Our synthetic training dataset included 150 ``how to/do'' variation queries, 1,421 paraphrased queries, and 952 singular/plural/article variation queries. We submit all query variations to the IR system to retrieve their corresponding contexts, which are then used to construct the final inputs. \par

% In addition to the 1,317 test samples used for evaluating the LLM's output accuracy, we created a separate test set to assess its consistency. We achieved this by applying the data synthesis methods described in Section \ref{sec:synthetic} to the original test set. This process yielded 176 "how to/do" variations, 912 paraphrased variations, and 491 singular/plural/article variations. These 1,579 synthetic queries, paired with their original queries and expected answers, form our test pairs for evaluating the LLM's output consistency. \par

Alongside the 1,317 test samples to measure accuracy, we created a test set to evaluate consistency using our data synthesis methods (Section \ref{sec:synthetic}). This produced 1,579 variations-176 ``how to/do'', 912 paraphrases, and 491 singular/plural/article changes-paired with original queries and expected answers for consistency testing. \par

\subsection{Metrics}
To assess the overall accuracy of the results of our RAG system, we employed the ROUGE-L \cite{lin-2004-rouge} and BLEU metrics with up to 4-grams \cite{papineni-etal-2002-bleu}, comparing the LLM-generated responses against the references provided. \par

To quantify the consistency of LLM response across input variations, we utilized three metrics: exact string match (EM), response similarity (RS) and Bert similarity (BS) measures. Given an original query $Q$ and its variant $Q'$, with $S$ and $S'$ representing the respective LLM responses, the exact string match is formally defined as: $$ \text{EM}(S, S') \iff S = S'.$$

For Response Similarity (RS), we determine semantic equivalence by thresholding the ROUGE score between the LLM's responses $S$ and $S'$: $$ RS(S, S') \iff \text{Rouge}(S, S') > T ,$$ where $T$ represents an empirically determined threshold used to ascertain whether two responses are considered semantically identical.

Furthermore, we define Bert Similarity (BS) between two LLMs responses $S$ and $S'$ to quantify the semantic similarity of them, as: $$ BS(S, S') \iff \text{Bert}(S, S'). $$

\subsection{Model Training and Merging}
Our experimental setup involved several distinct fine-tuning stages for the Llama-3.1-8B-Instruct model \cite{grattafiori2024llama3herdmodels} and Gemma-3-12B-Instruct model \cite{gemmateam2025gemma3technicalreport}. \par

We started by fine-tuning a baseline Llama-3.1-8B-Instruct and Gemma-3-12B-Instruct models for two epochs, using the original 1,421 training samples and only a cross-entropy loss function. \par

To investigate how triplet loss could boost LLM consistency, all subsequent fine-tuning experiments combined both cross-entropy loss and triplet loss, keeping the hyperparameters consistent with our initial baseline setup. \par

\begin{table*}[htp]
\centering
\fontsize{8.5pt}{10pt}\selectfont
\begin{threeparttable}
\caption{Comparison of Overall Accuracy and Consistency Metrics.} %Abbreviations: B = Baseline, SFT = Supervised Fine-tuned, TL = Triplet-loss, HTD = "How to/do" variation, SEM = Semantic variation, SPA = Singular/Plural/Article variation, ALL = All training data, Merged = Merged model.}
\begin{tabular}{l | c | c || c | c | c || c | c || c | c | c}
\hline
\multirow{2}{*}{} & \multicolumn{5}{c||}{Llama-3.1-8B-Instruct based LLMs} & \multicolumn{5}{c}{Gemma-3-12B-Instruct based LLMs} \\ \cline{2-11}
 & ROUGE & BLEU & EM & RS & BS & ROUGE & BLEU & EM & RS & BS \\ \hline\hline
B & 0.5123 & 0.2928 & 0.1051 & 0.2799 & 0.9246 & 0.4692 & 0.2338 & 0.0678 & 0.2609 & 0.9227 \\
B + SFT & 0.5208 & 0.3125 & 0.1482 & 0.3325 & 0.9266 & 0.5266 & 0.3297 & 0.2242 & 0.4009 & 0.9323 \\
B + SFT + TL & 0.5460 & 0.3460 & 0.1822 & 0.3530 & 0.9276 & 0.5206 & 0.3194 & 0.2331 & 0.4041 & 0.9337 \\
B + SFT + TL + HTD & \textbf{0.5493} & \textbf{0.3495} & 0.2250 & 0.3867 & 0.9264 & 0.5276 & 0.3255 & 0.2483 & 0.4364 & 0.9351 \\
B + SFT + TL + SEM & 0.5330 & 0.3339 & 0.2366 & 0.3965 & 0.9281 & 0.4966 & 0.3042 & 0.2673 & 0.4262 & 0.9314 \\
B + SFT + TL + SPA & 0.5364 & 0.3370 & 0.2111 & 0.3692 & 0.9262 & 0.5130 & 0.3170 & 0.2603 & 0.4231 & 0.9332 \\
B + SFT + TL + ALL & 0.5198 & 0.3230 & 0.2510 & 0.3986 & 0.9289 & 0.4879 & 0.2974 & \textbf{0.3382} & \textbf{0.4731} & 0.9357 \\
Merged & 0.5379 & 0.3380 & \textbf{0.2521} & \textbf{0.4129} & \textbf{0.9292} & \textbf{0.5356} & \textbf{0.3416} & 0.2932 & 0.4674 & \textbf{0.9373} \\ \hline
\end{tabular}
\label{tab:results}
    \begin{tablenotes}
    \small
    \item Abbreviations: B = Baseline (Llama-3.1-8B-Instruct or Gemma-3-12B-Instruct), SFT = Supervised Fine-tuned, TL = Triplet-loss, HTD = ``How to/do'' variation, SEM = Semantic variation, SPA = Singular/Plural/Article variation, ALL = All training data, Merged = Merged model.
    \end{tablenotes}
\end{threeparttable}
\end{table*}

Following this, we fine-tuned five distinct Llama-3.1-8B-Instruct LLMs. One was fine-tuned on our base training set exclusively. The other three were fine-tuned on this base set, each augmented with a specific synthetic data type: 176 ``how to/do'' variations, 912 paraphrased samples, or 491 singular/plural/article variations (more details on this in Section \ref{subsec:triplet}). The final model was fine-tuned using all available training data combined. Finally, we merged these three individually fine-tuned LLMs using the methodology described in Section \ref{subsec:merging}. We repeated the same fine-tuning and merging steps for the Gemma-3-12B-Instruct LLMs to ensure consistent evaluation across model's architectures. \par

%%Before fine-tuning, we normalized all queries to the LLM were normalized through lower casing and punctuation removal. \par

All fine-tuned models, including the Llama-3.1-8B-Instruct and Gemma-3-12B-Instruct baselines, were comprehensively evaluated in two dedicated test sets designed to assess both accuracy and consistency measures, as described in Section \ref{subsec:datasets}. We present complete experimental results in Table \ref{tab:results}.

\subsection{Results}
In Table \ref{tab:variation_stat}, we present four types of query variations that lead to response inconsistency. Table \ref{tab:results} quantitatively shows that the baseline model (Llama-3.1-8B-Instruct) achieves moderate overlap with human references (ROUGE: 0.5123, BLEU: 0.2928) but demonstrates the lowest consistency (EM: 0.1051, RS: 0.2799, BS: 0.9246). This demonstrates the model often fails to generate consistent responses to semantically equivalent queries.

The fine-tuned model, as shown in Table \ref{tab:results}, demonstrates a modest improvement over the baseline, w.r.t accuracy and consistency. While it yields somewhat better text overlap and initial gains in consistency, its performance, particularly in EM, RS and BS, suggests that general fine-tuning provides only limited progress towards truly consistent response for varied inputs.

Incorporating triplet loss significantly boosts performance across all metrics, as seen by comparing the triplet-loss model to the fine-tuned model in Table \ref{tab:results}. The triplet-loss model achieved higher ROUGE (0.5460) and BLEU (0.3460) scores, indicating better content and lexical alignment. The model shows improvement in consistency, the EM score dramatically improved by 73.4\% to 0.1822 (from 0.1051), while RS score saw a substantial 26.1\% increase to 0.3530 (from 0.2799). These results  underscore the effectiveness of integrating triplet loss in fine-tuning strategies for LLMs, leading to significantly more robust and consistent response generation.

As shown in the Table \ref{tab:results}, individual variation models—specifically the \emph{How to/Do},\emph{Semantic}, and \emph{Singular/Plural/Article} variation models—consistently outperform the baseline in both accuracy and consistency. This demonstrates the effectiveness of specialized fine-tuning with synthetically generated data\par

Surprisingly, models fine-tuned on individual synthetic datasets outperformed the combined-data model in accuracy. However, the combined model achieved higher consistency, suggesting that merging diverse variation types may introduce conflicting signals or biases that impact accuracy. 
\par

%In terms of consistency metrics, the merged model achieves the highest scores for both EM at 0.2521 and RS at 0.4129. This performance significantly surpasses all other models. For EM, it represents an impressive 139.87\% improvement over the baseline model and still leads the next best individual model (Semantic variation model at 0.2366) by approximately 6.55\%. Similarly, its RS score is a 47.52\% improvement over the baseline and approximately 4.14\% higher than the semantic variation model (0.3965). This indicates that the merging strategy is highly effective in ensuring LLM responses are more reliably identical or semantically equivalent even when faced with varied inputs. \par
The merged model consistently delivers the most robust and balanced performance across all metrics, with notable strength in response consistency. In terms of consistency metrics, the merged model achieves the highest scores for both EM at 0.2521, RS at 0.4129 and BS at 0.9292. This performance significantly surpasses all other models. For EM, it represents an impressive 139.87\% improvement over the baseline model and still leads the next best model (``B + SFT + TL + ALL'' model at 0.2510) by approximately 0.44\%. Similarly, its RS score is a ~47.52\% improvement over the baseline and approximately ~3.59\% higher than the second best model. This indicates that the merging strategy is highly effective in ensuring LLM responses are more reliably identical or semantically equivalent even when faced with varied inputs.

Regarding accuracy-based metrics, while the merged model's ROUGE score (0.5379) and BLEU score (0.3380) are marginally lower than the top performer (the ``B + SFT + TL + HTD'' with ROUGE 0.5493 and BLEU 0.3495), they are still very good and highly competitive. This demonstrates that the merging process successfully enhances consistency without a significant trade-off in overall accuracy or fluency. The merged model effectively combines the strengths of its constituent specialized models, making it the most well-rounded and high-performing solution for both accurate and consistent RAG system responses. \par

Table \ref{tab:results} also reports accuracy and consistency metrics for Gemma-3-12B-Instruct models. Overall, the trend of model improvements closely mirrors that of the Llama-3.1-8B-Instruct experiments: baseline models exhibit moderate ROUGE and BLEU scores with the lowest consistency (EM: 0.0678, RS: 0.2609, BS: 0.9227), fine-tuning improves both accuracy and consistency, incorporating triplet loss further boosts response reliability, and models fine-tuned on individual synthetic variations outperform the baseline in both accuracy and consistency. For Gemma, the ``B + SFT + TL + ALL'' model achieves the highest consistency metrics (EM: 0.3382, RS: 0.4731), similar to the trend observed for Llama, where combined-data models also prioritize consistency over raw accuracy. The merging strategy consistently delivers the most robust and balanced performance across all metrics.

Key differences are notable, however. The Gemma baseline shows lower initial accuracy and EM than the Llama baseline, suggesting a larger model does not automatically guarantee consistent responses.  The merged Gemma model attains the highest ROUGE (0.5356), BLEU (0.3416), and BS (0.9373), slightly outperforming Llama’s merged model on accuracy and semantic similarity, though EM is slightly lower than Gemma’s ``B + SFT + TL + ALL'' model, indicating a minor trade-off in exact match consistency.

Overall, while the pattern of improvement, baseline → fine-tuned → triplet-loss → specialized variation → merged, is consistent across both model families, the larger Gemma-3-12B-Instruct benefits more from combined data fine-tuning, achieving higher accuracy and semantic similarity, while the merging strategy ensures robust and balanced performance similar to both Llama and Gemma models.

\section{Conclusion}
In this work, we identify four types of semantically insignificant query variations that cause inconsistent LLM responses. We quantify response similarity and show that baseline models and standard fine-tuning exhibit low consistency. To address this, we propose a novel approach combining synthetic data generation, Triplet Loss training, and layer-wise model merging guided by consistency-oriented weights. \par

 Our experiments show the merged model significantly outperforms baselines and specialized models, achieving superior Exact Match and Response Similarity scores, thus demonstrating enhanced consistency while maintaining strong accuracy. This work presents a compelling pathway towards more trustworthy LLMs and opens avenues for future research, including adaptive merging, expanded consistency definitions, and the application of this method to diverse public datasets.  We also plan to construct and publicly release benchmarks that mimic the identified query variations to further evaluate and demonstrate the effectiveness of our approach. Additional future directions include addressing inconsistency cases arising from retrievers, which were beyond the scope of this study.
\section{Limitations}
While the proposed work has been evaluated on industrial data, there is scope to create public benchmark and evaluate the method.  We will explore creating creating benchmarks for evaluating consistency in responses due to variations in the query.  \par
The work is limited to variations in the query where the retrievers results don't significantly change. This can explored as future direction for research.  \par
Finally, we experimented with 2 Large Language Models with optimal settings for fine-tuning. There is scope to explore additional hyper parameter configurations.  

\bibliography{ref}

\end{document}